\documentclass[sigconf]{acmart}

\copyrightyear{2021}
\acmYear{2021}
\setcopyright{acmlicensed}\acmConference[CIKM '21]{Proceedings of the 30th ACM International Conference on Information and Knowledge Management}{November 1--5, 2021}{Virtual Event, QLD, Australia}
\acmBooktitle{Proceedings of the 30th ACM International Conference on Information and Knowledge Management (CIKM '21), November 1--5, 2021, Virtual Event, QLD, Australia}
\acmPrice{15.00}
\acmDOI{10.1145/3459637.3482279}
\acmISBN{978-1-4503-8446-9/21/11}

\usepackage{import}
\usepackage{bm}
\usepackage{diagbox}
\usepackage{multirow}
\usepackage{graphicx}
\usepackage{subcaption}
\usepackage{mwe}

\newcommand{\R}{\mathbb{R}}
\AtBeginDocument{%
  \providecommand\BibTeX{{%
    \normalfont B\kern-0.5em{\scshape i\kern-0.25em b}\kern-0.8em\TeX}}}

\begin{document}
\fancyhead{}

\title{ Multi-Relational Graph based Heterogeneous Multi-Task Learning in Community Question Answering}

\author{Zizheng Lin$^{1}$, Haowen Ke$^{1}$, Ngo-Yin	Wong$^{1}$, Jiaxin	Bai$^{1}$, Yangqiu Song$^{1}$, Huan Zhao$^{2}$,  Junpeng Ye$^{3}$}
\affiliation{%
  \institution{$^{1}$Department of Computer Science and Engineering, HKUST, Hong Kong, China\\
  $^{2}$4Paradigm Inc., Beijing, China\\
  $^{3}$Tencent Technology (SZ) Co., Ltd., Shenzhen, China}
  }
\email{{zlinai,hkeaa,nywongac,jbai,yqsong}@cse.ust.hk, zhaohuan@4paradigm.com, jayjpye@tencent.com}


%\keywords{datasets, neural networks, gaze detection, text tagging}

\begin{comment}
\begin{teaserfigure}
  \includegraphics[width=\textwidth]{sampleteaser}
  \caption{Seattle Mariners at Spring Training, 2010.}
  \Description{Enjoying the baseball game from the third-base
  seats. Ichiro Suzuki preparing to bat.}
  \label{fig:teaser}
\end{teaserfigure}
\end{comment}

\begin{abstract}
Various data mining tasks have been proposed to study Community Question Answering (CQA) platforms like Stack Overflow. The relatedness between some of these tasks provides useful learning signals to each other via  Multi-Task Learning (MTL). However, due to the high heterogeneity of these tasks, few existing works manage to jointly solve them in a unified framework. To tackle this challenge, we develop a multi-relational graph based MTL model called Heterogeneous Multi-Task Graph Isomorphism Network (HMTGIN) which efficiently solves heterogeneous CQA tasks. 
In each training forward pass, HMTGIN embeds the input CQA forum graph by an extension of Graph Isomorphism Network and skip connections. The embeddings are then shared across all task-specific output layers to compute respective losses. Moreover, two cross-task constraints based on the domain knowledge about tasks' relationships are used to regularize the joint learning. 
In the evaluation, the embeddings are shared among different task-specific output layers to make corresponding predictions.
To the best of our knowledge, HMTGIN is the first MTL model capable of tackling CQA tasks from the aspect of multi-relational graphs.
To evaluate HMTGIN's effectiveness, we build a novel large-scale multi-relational graph CQA dataset with over two million nodes from Stack Overflow. Extensive experiments show that: $(1)$ HMTGIN is superior to all baselines on five tasks; $(2)$ The proposed MTL strategy and cross-task constraints have substantial advantages.
\end{abstract}
\maketitle

\section{Introduction} 
Community Question Answering (CQA) forums like Stack Overflow\footnote{https://stackoverflow.com/} help millions of users to seek solutions or share their knowledge.
Various CQA tasks like duplicate question detection and answer recommendation have been extensively studied \cite{srba2016comprehensive}.
Due to rich interconnections between components of a CQA platform, some CQA tasks are related to each other.
For example, tag recommendation can be solved using a link prediction model and the semantic representation of keywords reflected by this task can also help evaluate a similar link prediction model in duplicate question detection.
Moreover, duplicate question detection would also help personalize answer recommendation as they both evaluate text similarities, although the latter should be formulated as a ranking problem.
Thus, it is natural to use Multi-Task Learning (MTL)  \cite{caruanamulti} to jointly solve different CQA tasks to obtain better overall performance.

However, most existing MTL frameworks only consider similar tasks \cite{zhang2017survey,ruder2017overview} like considering classification and regression of the same learning features \cite{YangKX09}, or assuming that the features are different but label spaces are the same across tasks \cite{JinZPDLH15}. Regarding heterogeneous tasks (i.e., tasks with very different properties such as contrasting objective functions and learning features), these MTL frameworks cannot be effective.
In the case of tackling CQA tasks, there are two major challenges for heterogeneous MTL (e.g., jointly solve link prediction, classification, and ranking over different types of nodes). 

First, it is non-trivial to share features for heterogeneous CQA data in the semantic space. Specifically, different entities, like users and answers barely share similar semantic space, as they are usually connected through complicated paths in different types of relations.
It may be not sufficient for the algorithm to learn the feature dependency, which is crucial for effectively sharing features across tasks, directly from training feature representations.
Thus, a multi-relational graph based model should be considered to build relationships among entities in a CQA platform.

Second, there are explicit relationships among output labels in the label space. However, to our best knowledge, existing MTL algorithms do not explicitly model the label relations.
For example, in CQA, a more reputable user's answer is more likely to be ranked higher, but the relation of preferences cannot be explicitly reflected by the corresponding representations of users and answers. 
Thus, besides data-driven learning, explicit label-based constraints can be imposed to regularize the representation learning across different CQA tasks.

To address the aforementioned challenges, we propose a novel MTL model named Heterogeneous Multi-Task Graph Isomorphism Network (HMTGIN) that efficiently learns multiple heterogeneous CQA tasks on any given CQA forum graph, despite the possibly substantial heterogeneity of the tasks and graph. HMTGIN adopts the hard parameter sharing mechanism \cite{ruder2017overview} to improve efficiency and reduce memory consumption. Inside HMTGIN, we design the Multi-Relational Graph Isomorphism Network (MRGIN), a multi-relational variant of the Graph Isomorphism Network (GIN) \cite{xu2018powerful}, and employ it with skip connections \cite{he2016deep} to learn the node embeddings shared across tasks. Besides learning from data, HMTGIN also imposes cross-task constraints on tasks' relationships to capture the feature and label dependency across heterogeneous tasks. 

Despite some existing MTL works \cite{yang2019advanced,yang2019knowledge} or graph mining \cite{akoglu2012opavion} on CQA, to the best of our knowledge, no previous MTL algorithm tackles CQA from the aspect of multi-relational graphs. 

To evaluate the effectiveness of the proposed approach, we build a novel large-scale multi-relational graph CQA dataset with over two million nodes from Stack Overflow. We then define five different CQA tasks which can be categorized into link prediction, ranking, and classification problems. Extensive experiments are conducted to compare HMTGIN's performance with corresponding baselines in each task and to examine the effect of the proposed MTL mechanism and cross-task constraints. 
%Experiments are only carried out on the constructed Stack Overflow dataset because our approach can be directly applied to other CQA platforms with trivial modifications.

%Contributions
Our main contributions are as follows: 

$\bullet$ We propose a novel MTL model termed HMTGIN that efficiently mines any given CQA forum graph, where the graph and tasks can have immense heterogeneity. To the best of our knowledge, HMTGIN is the first MTL model that tackles CQA tasks from the perspective of multi-relational graphs.

$\bullet$ We construct a novel million-scale multi-relational graph CQA dataset from Stack Overflow.

$\bullet$ We perform extensive experiments of five tasks, where HMTGIN is shown to be superior to all baselines. Further empirical analysis demonstrates the considerable improvements of our MTL strategy and cross-task constraints.

Our dataset and code are publicly available at 
\url{https://github.com/HKUST-KnowComp/HMTGIN}.

\section{Related Work}

We discuss the related work in four-fold. 

\subsection{Community Question Answering (CQA)}

%Overview
CQA platforms enable people to seek and share knowledge effectively. For instance, Stack Overflow is a prominent CQA platform about programming, where many developers actively learn from others or share their expertise.
%Common tasks
Various CQA tasks have been extensively studied \cite{srba2016comprehensive}, like post recommendation \cite{xu2012dual} and duplicate post detection \cite{wang2020duplicate}. In this paper, we study five CQA tasks of the following three types: link prediction, ranking, and classification. Details of these tasks are in Section \ref{sec:task_description}.
%Limitation and corresponding advantages
Despite some existing works about MTL for CQA \cite{yang2019advanced,yang2019knowledge}, to the best of our knowledge, no previous MTL algorithm can solve CQA tasks from the aspect of multi-relational graphs, and the tasks considered in previous MTL works are relatively homogeneous. In contrast, our approach is the first multi-relational graph based MTL model tackling highly heterogeneous CQA tasks.

\subsection{Multi-Task Learning (MTL)}

%Overview
MTL aims to strengthen a model's generalizability by jointly learning multiple related tasks \cite{caruanamulti,zhang2017survey,ruder2017overview}. Successful applications in many domains like Natural Language Processing \cite{ruder2019latent,liu-etal-2017-adversarial,sogaard2016deep} have verified MTL's effectiveness.
Recent MTL models usually employ deep neural networks  \cite{ruder2019latent,liu-etal-2017-adversarial}. Most of them adopt either hard parameter sharing where each task shares the same hidden layers while keeping its output layer, or soft parameter sharing where every task has its model \cite{ruder2017overview}. 
%Limitation
However, most existing MTL models only undertake tasks with small heterogeneity \cite{zhang2017survey,ruder2017overview},  like considering regression and classification of identical learning features \cite{YangKX09}, or assuming that features are different but label spaces are the same \cite{JinZPDLH15}. 
Moreover, many existing MTL algorithms are purely data-driven \cite{zhang2017survey,ruder2017overview,li2020multi},  which usually prevents them from capturing the label dependency across tasks due to the large heterogeneity of multi-tasks.
%Corresponding advantages of proposed work
In contrast, our MTL model efficiently performs highly heterogeneous learning tasks, where two cross-task constraints are imposed on tasks' relationships to regularize the joint learning.

\subsection{Graph Neural Networks}
%Overview
Graph Neural Networks (GNNs) typically perform graph representation learning by recursive neighborhood aggregation \cite{9046288}. Many GNN variants \cite{kipf2017semi,xu2018representation} led to significant advancement in various graph mining tasks like link prediction.
%GIN
Additionally, \cite{xu2018powerful} conducted a systematic study on GNNs' representational power via the connections between GNNs and Weisfeiler-Lehman (WL) graph isomorphism test \cite{weisfeiler1968reduction}. Under this framework, \cite{xu2018powerful} proposes Graph Isomorphism Network (GIN) which provably accomplishes the utmost discriminative power among all kinds of GNNs.
%multi-relational GNN
Different from the above GNNs which only consider homogeneous graphs, several variants \cite{schlichtkrull2018modeling, WangJSWYCY19,HuDWS20,CenZZYZ019,abs-2004-00216,zhao2017meta,xiao2019beyond} have been designed for multi-relational graphs.
%Limitation
Although GNNs have demonstrated superb learning abilities, few works adopt GNNs in MTL models \cite{wang2020m2grl,WangXGL0CYWC20}.
%Corresponding advantages of proposed work
Hence, we devise an extension of GIN termed Multi-Relational GIN as one of the main components of our MTL model to facilitate multi-relational graph representation learning. Besides GIN, our framework can be easily extended to incorporate other GNN architectures to further boost performance.

\subsection{Constraint Learning}

%Overview
Constraint learning aims to improve a model's performance by incorporating domain knowledge as constraints.
%Must-link and cannot-link 
One popular constraint learning framework is the cannot-link and must-link modeling, where constraints are typically based on ground-truth labels \cite{basu2004probabilistic}. 
%Various tasks like document clustering \cite{song2012constrained,wang2015incorporating} have adopted the constraints under this framework as indirect supervision. 
%CCM and PR
Another powerful scheme is the Constrained Conditional Models (CCM) \cite{chang2012structured} where learning is separated from the knowledge-aware inference. Furthermore, Posterior Regularization (PR) \cite{ganchev2010posterior} embodies knowledge by a joint learning and inference method. 
%CCM and PR have been extensively used in sundry tasks like relation extraction \cite{chan2010exploiting,chen2011domain}.
%Limitation and corresponding advantages
However, few works apply constraints on tasks' relationships to enhance MTL. Thus, in our MTL model, we impose cross-task constraints on tasks' relationships, which is shown to be effective.

\section{Preliminaries}\label{sec:preliminaries}

\begin{figure}[t]
    \centering
    \resizebox{\columnwidth}{130pt}{%
    \includegraphics{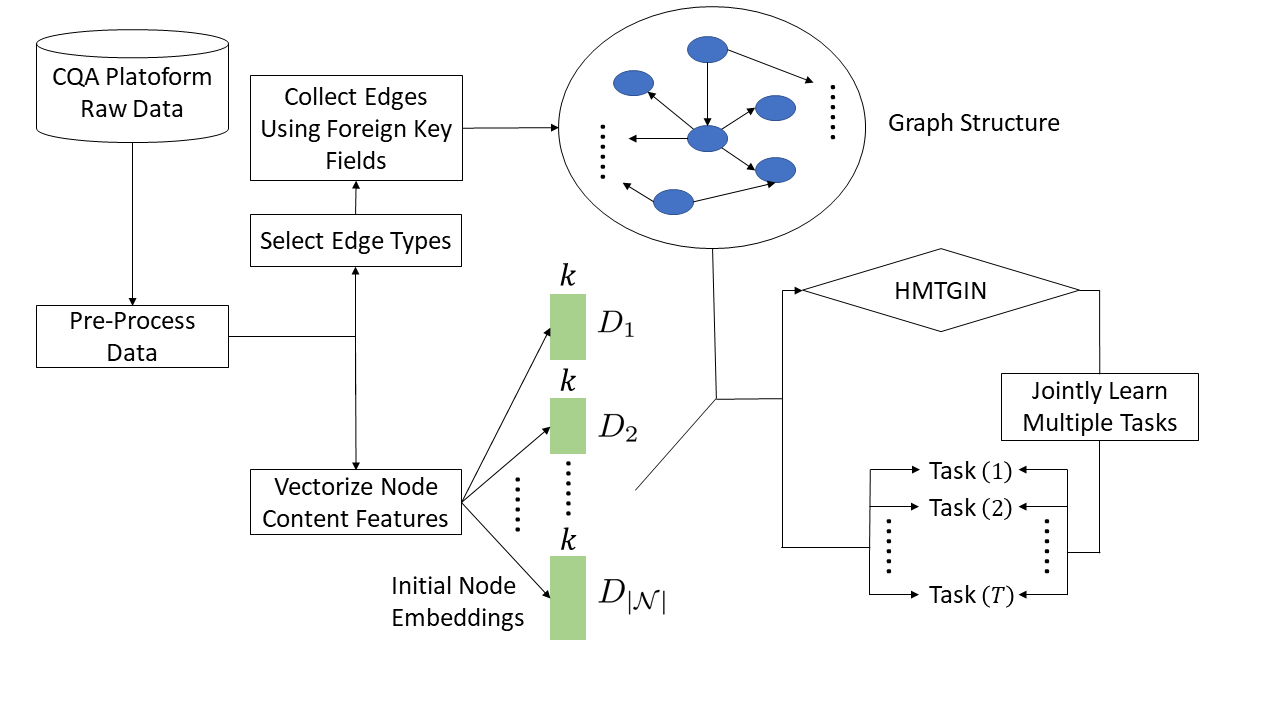}
    }
    \caption{Overview of the generic framework. $k$ is the dimensions of initial node embeddings, and $\mathcal{N}$ is the set of all node types, and $D_i, (i = 1,2, ..., |\mathcal{N}|)$ denote cardinalities of respective node types, and $T$ is the number of tasks.}
    \label{fig:pipeline}
\end{figure}

In this section, we describe the tasks and frequently used notations.
Figure \ref{fig:pipeline} depicts the generic framework for building the multi-relational graph from a CQA raw dataset, and how HMTGIN performs MTL on the built dataset. 

\subsection{Tasks Description}\label{sec:task_description}

We divide all tasks into the following three types: link prediction, ranking, and classification.

\subsubsection{Link Prediction}

Link prediction predicts whether an edge exists between a given node pair, where some of the existing edges are hidden from the input graph. Regarding CQA, link prediction can benefit various applications like recommendations. Here are the two link prediction tasks: $(1)$ Tag recommendation: given a question-tag pair, predict whether the tag belongs to that question; $(2)$ Duplicate question detection: given a question-question, predict whether there is a `Duplicate' type of edge.
To obtain negative examples, we randomly sample twice as many node pairs as the existing links for each task, where no target link exists between any of the sampled node pairs.

\subsubsection{Ranking}

A ranking task obtains sorts an input item list for a user to quickly identify the desired item(s). Since many CQA platform questions have many associated answers, generating a personalized ranking of the answers can greatly save users' time. Hence, we define a task called answer recommendation as follows: given a question with at least eight answers, provide a ranking of all its answers such that the accepted answer has a high rank. Each sample contains a question index, a list its associated answers' indices, and the list positional index of the accepted answer as the label.

\subsubsection{Classification}\label{sec:clf}

A classification task categorizes a certain node attribute. Since some attributes like questions' \textit{score} attribute might reveal crucial node properties, categorization of them may help identify important nodes. Here are the two link prediction tasks: $(1)$ Answer score classification: classify an answer score into one of the integers in $[0,3]$, where a higher value means higher \textit{score} attribute; $(2)$ User reputation classification: classify a user's reputation into one of the integers in $[0,4]$, where a higher value indicates higher \textit{reputation} attribute.
During the data pre-processing, labels for each target attribute are generated in advance by dividing the values into different intervals, and the corresponding attribute is masked in the dataset.

\subsection{Notations}\label{sec:notation}

Following notations are used throughout the remaining paper: the directed and labeled input multi-relational graph is denoted as $\mathcal{G}=(\mathcal{V},\mathcal{E})$ with a node type mapping function $\phi: \mathcal{V} \rightarrow \mathcal{N}$, and edge type mapping function $\psi: \mathcal{E} \rightarrow \mathcal{R}$, where each node $v \in \mathcal{V}$ has one particular node type $\phi(v) \in \mathcal{N}$, and each edge $e \in \mathcal{E}$ has one particular edge type $\psi(e) \in \mathcal{R}$, and $\mathcal{R}$ does not include the self-loop edge type.
Additionally, we use Multi-Layer Perceptron (MLP) or $MLP$ for a feedforward neural network with \textit{zero or more layers}. Moreover, we use $\sigma$ for activation functions like RELU. In addition, we use $BN$ for Batch Normalization (BN) \cite{ioffe2015batch}. Furthermore, we use $f$ for sigmoid function.
We also use $\circ$ for constructing the network by stacking different layers.

\section{Methodology}\label{sec:methodology}

We first present MRGIN. Then we explain all task-specific output layers. Thereafter, we explain the cross-task constraints on tasks' relationships. Finally, we describe the whole MTL algorithm.

\subsection{Multi-Relational Graph Isomorphism Network (MRGIN)}\label{sec:MRGIN}

We propose MRGIN which, together with skip connections \cite{he2016deep}, embeds the input multi-relational graph. MRGIN is inspired by GIN \cite{xu2018powerful} that has been theoretically shown to have the maximum discriminative power among all types of GNNs. Specifically, let $d^{(l')}$ be the dimension of $(l')$-th MRGIN layer's node representation. The representation of a node $v_i \in \mathcal{V}$ in $(l+1)$-th MRGIN layer, denoted as $\mathbf{h_i}^{(l+1)} \in \R^{d^{l+1}}$, is computed as:
\begin{multline}\label{eq:MRGIN}
     \mathbf{h_i}^{(l+1)} = \sigma^{(l+1)} \circ BN^{(l+1)} \circ MLP^{(l+1)} \circ 
     \\
     \sigma^{(l+1)} (\sum_{r\in \mathcal{R}} \sum_{j\in \mathcal{N}_i^r } \mathbf{W_r}^{(l+1)}\mathbf{h_j}^{(l)} 
     + (1+\epsilon^{(l+1)}) \mathbf{W_0}^{(l+1)} \mathbf{h_i}^{(l)}) ,
\end{multline}
where $\mathcal{N}_i^r$ is the set of neighbor indices of node $v_i$ under edge type $r \in \mathcal{R}$, and $\mathbf{h_j}^{(l)} \in \R^{d^{(l)}}$ as well as $\mathbf{h_i}^{(l)} \in \R^{d^{(l)}}$ are the representations of nodes $v_j$ and $v_i$ respectively in $l$-th MRGIN layer, and both $\mathbf{W_r}^{(l+1)} \in \R^{d^{(l+1)} \times d^{(l)}}$ and $\mathbf{W_0}^{(l+1)} \in \R^{d^{(l+1)} \times d^{(l)}}$ are type-specific transformation matrices, and $\epsilon^{(l+1)}$ is a scalar which is either trainable or fixed. Updating one layer is to concurrently evaluate Equation (\ref{eq:MRGIN}) for all nodes. 

Briefly speaking, MRGIN aggregates transformed neighboring nodes' representations by sum pooling. Moreover, every node's feature vector is scaled by $1+\epsilon$ before being aggregated into its representation in the next layer. The vector produced by the summation is then processed via a non-linear activation function, followed by an MLP, a BN, and finally another non-linear activation function. One crucial distinction between MRGIN and GIN is that the transformation matrices in the aggregation of MRGIN depend on edges' orientations and types, which enables MRGIN to exploit the input graph's structural and relational information.

\subsection{Task-Specific Output Layers (TSOLs)}\label{sec:tsol}

We design TSOLs for task types in Section \ref{sec:task_description} as follows:

\subsubsection{Link Prediction} 

We associate each link prediction task $t$ with a diagonal matrix $\mathbf{D_t} \in \R^{k \times k}$ initialized by a standard normal distribution, where $k$ is the dimension of the input node embeddings for TSOLs.
Given a node pair $(v_i,v_j)$ for a task $t$,  we follow the DisMult algorithm \cite{yang2014embedding} to calculate the link prediction score $S_{ij}$ as follows:
\begin{align}
    S_{ij} = \mathbf{h_i^{'T}} \mathbf{D_t} \mathbf{h_j^{'}} ,
\end{align}
where $\mathbf{h_i^{'}}$ and $\mathbf{h_j^{'}} \in \R^{k}$ is the input node embeddings for TSOLs of $v_i$ and $v_j \in \mathcal{V}$ respectively.
Suppose the type of the target link in task $t$ is $r_t$, define a set of node pairs as follows:
\begin{multline}
\mathcal{H}_t = \{ \, (v_i, v_j) \, | \, v_i \in \mathcal{V}, v_j \in \mathcal{V}, \text{ $\exists \, e \in$ $E$ from $v_i$ to $v_j$ such that} \\ \psi (e) = r_t \, \} .
\end{multline}
Let $\mathcal{H}_t^{'}$ be the set of node pairs obtained by the negative sampling procedure mentioned in Section \ref{sec:task_description} for task $t$. Then task $t$'s loss $L_t$ is:
\begin{multline}\label{eq:link_pred_task_loss}
    L_t = - \frac{1}{|\mathcal{H}_t| + |\mathcal{H}_t^{'}|} \sum_{(v_i, v_j) \, \in \, \mathcal{H}_t \cup \mathcal{H}_t^{'}} [w \cdot y_{ij} \cdot \log(f(S_{ij})) + \\ (1-y_{ij}) \cdot \log(1-f(S_{ij}))], 
\end{multline}
where $w$ is the weight of positive samples (i.e., node pairs in $H_t$), and $y_{ij} = 1$ if $(v_i, v_j) \in \mathcal{H}_t$ and $y_{ij} = 0$ if $(v_i, v_j) \in \mathcal{H}_t^{'}$.

\subsubsection{Ranking}

Regarding ranking task $t$, the parameters of its TSOL include $\mathbf{w_{t}} \in \R^{2k}$ and $b_t \in \R$. Denote the $i$-th input sample as $(Q_i, \mathcal{A}(Q_i))$ where $Q_i$ is a question index and $\mathcal{A}(Q_i)$ is a list containing the indices of the question's answers. Denote the sample's label as $y_i$. Motivated by RankNet \cite{burges2005learning}, we compute the loss term of $i$-th sample, denoted as $L_{t}^{i}$, as follows:
$(1)$ Retrieve the input node embeddings for TSOLs by $Q_i$ and $\mathcal{A}(Q_i)$; $(2)$ Concatenate the question embedding with each of the answer embeddings. Denote the concatenated embeddings as $\mathbf{M_t^{i}} \in \R ^{ 2k \times |\mathcal{A}(Q_i)| }$; $(3)$ Compute the ranking score of answer in the $j$-th position of $\mathcal{A}(Q_i)$, denoted as $R_j$, by 
    \begin{align}\label{eq:ranking score}
        R_j = \sigma(\mathbf{w_{t}^{T}} \mathbf{M_t^{i}}[:,j] + b_t) ,
    \end{align}
where $\mathbf{M_t^{i}}[:,j]$ is the $j$-th column of $\mathbf{M_t^{i}}$ matrix; $(4)$ Calculate the loss term as 
    \begin{align}\label{eq:loss term}
        L_{t}^{i} = - \frac{1}{|\mathcal{A}(Q_i)|} \sum_{j=1, j \neq y_i}^{|\mathcal{A}(Q_i)|} [\log(f(R_{y_i} - R_j))] .
    \end{align}
The loss of this task denoted as $L_t$ is the average of all input samples' loss terms.
Since Equation (\ref{eq:loss term}) is essentially a pairwise ranking loss, minimizing $L_t$ amounts to maximizing the difference between the ranking score of the accepted answer and those of the other answers in each input sample, which results in recommendations where the ranks of the accepted answers tend to be high. 

\subsubsection{Classification}

Regarding a classification task $t$, we use a MLP with one layer as its TSOL. The weight and bias of this MLP are $\mathbf{W_t} \in \R^{K_t \times k}$ and $\mathbf{b_t} \in \R^{K_t}$ respectively, where $K_t$ is the number of possible classes in task $t$. Given an input sample, which is a node embedding vector $\mathbf{h_i^{'}} \in \R^{k}$, the TSOL first computes the logit vector, denoted as $\mathbf{x_i} \in \R^{K_t}$, by a linear transformation:
\begin{align}\label{eq:clf_logit}
    \mathbf{x_i} = \mathbf{W_t} \mathbf{h_i^{'}} + \mathbf{b_t} .
\end{align}
Then the loss of task $t$, denoted as $L_t$, is calculated by softmax cross-entropy:
\begin{align}
    L_t = - \frac{1}{N^{'}} \sum_{i=1}^{N^{'}} [\log(\frac{\exp(\mathbf{x}[y_i])}{\sum_{j=1}^{K_t} \exp(\mathbf{x}[j])})] ,
\end{align}
where $N^{'}$ is the number of input samples and $y_i$ is the class label of the $i$-th sample.

\subsection{Cross-task Constraints}\label{sec:cross-task constraint} 
Besides data-driven learning, our model can be equipped with domain knowledge about tasks' relationships via incorporating several cross-task constraints into the objective function to regularize the joint learning. 
In this paper, we design two cross-task constraints as a demonstration of our constraint learning framework. Apart from these constraints, our framework can easily incorporate more cross-task constraints exploiting the domain knowledge about tasks' relationships by simply adding extra constraint loss functions into the objective.

Each of the two cross-task constraints relies on one of the following two assumptions based on the domain knowledge about CQA. Given two answers $A_1$ and $A_2$ of a question: 
\begin{enumerate}
    \item {\bf Assumption 1}:  if $A_1$'s \textit{score} attribute is higher than that of $A_2$, then $A_1$ is more likely to be accepted;
    \item {\bf Assumption 2}: if $A_1$'s owner's \textit{reputation} attribute is higher than that of $A_2$, then $A_1$ is more likely to be accepted.
\end{enumerate}
Assumption $1$ is reasonable as an answer with higher \textit{score} attribute usually has higher quality. Similarly, Assumption $2$ is also legitimate since a user with higher \textit{reputation} attribute typically has higher capability, which implies that the user's answer tends to be better. Hence, we propose the following two cross-task constraints. 
\begin{figure*}
    \centering
    \begin{subfigure}[b]{0.475\textwidth}
            \centering
            \includegraphics[width=\textwidth]{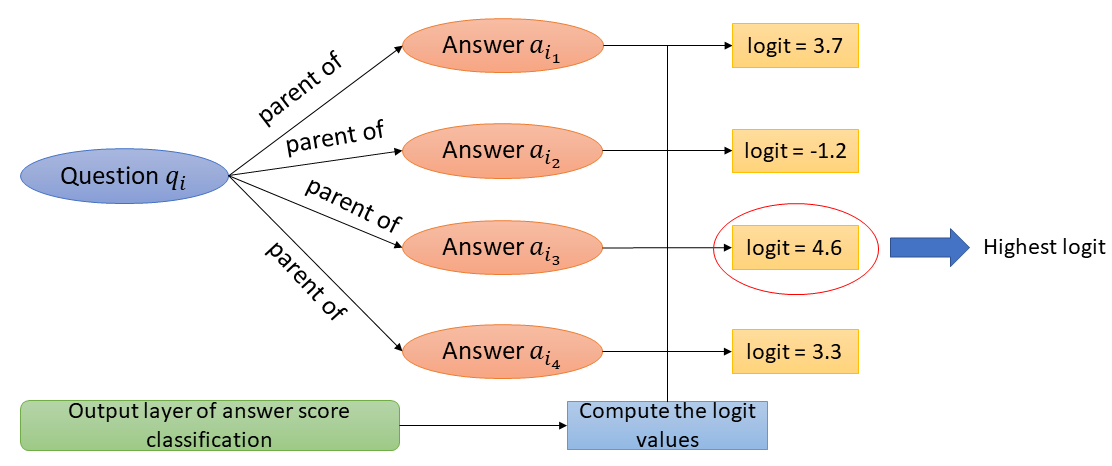}
            \caption{Constraint 1 step (1): compute logits via answer classification output layer and select the answer with highest logit.}
            \label{fig:constraint_1_example_1}
    \end{subfigure}
    \hfill
    \begin{subfigure}[b]{0.475\textwidth}
            \centering
            \includegraphics[width=\textwidth]{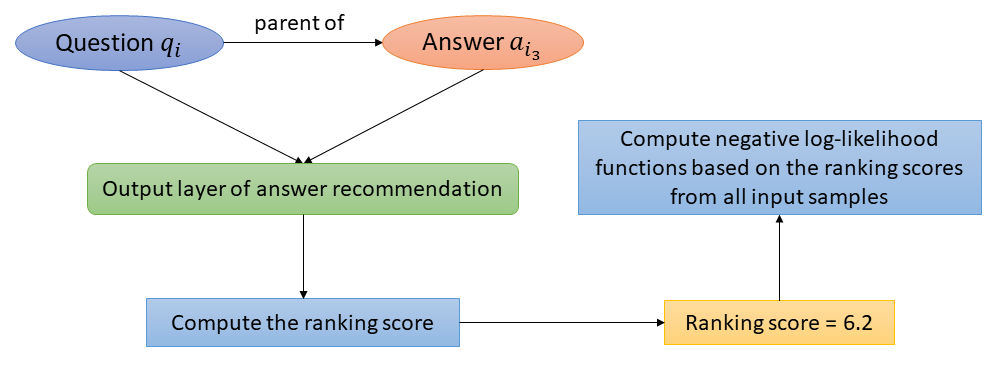}
            \caption{Constraint 1 step (2), (3): compute the ranking score(s) of answer(s) with the highest logit, then compute negative log-likelihoods which are averaged to get the overall constraint loss.}
            \label{fig:constraint_1_example_2}
    \end{subfigure}
    \vskip\baselineskip
    \begin{subfigure}[b]{0.475\textwidth}
            \centering
            \includegraphics[width=\textwidth]{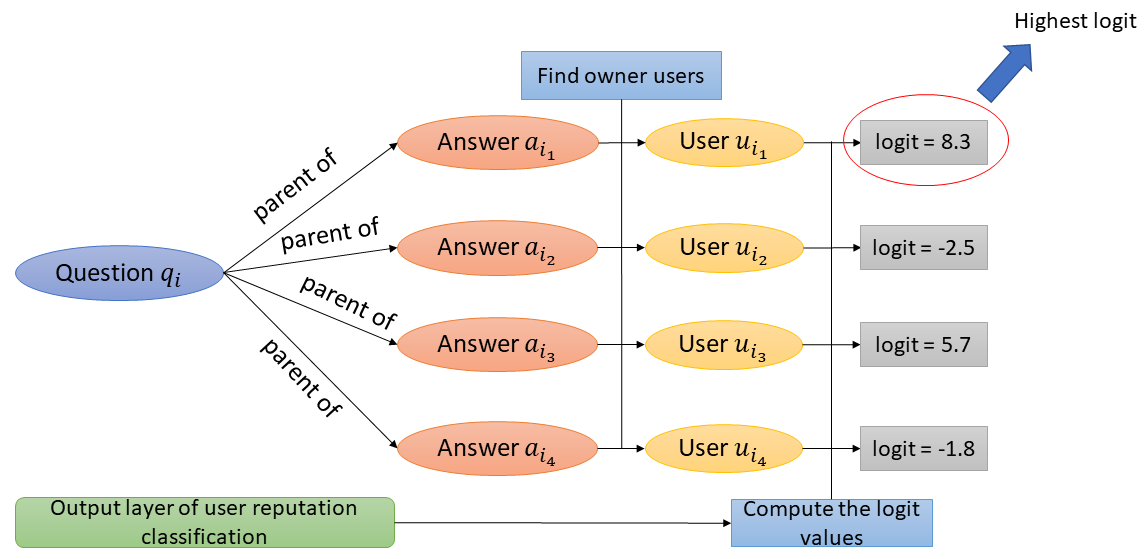}
            \caption{Constraint 2 step (1), (2): find owner users and compute their logits via user classification output layer.}
            \label{fig:constraint_2_example_1}
    \end{subfigure}
     \hfill
     \begin{subfigure}[b]{0.475\textwidth}
            \centering
            \includegraphics[width=\textwidth]{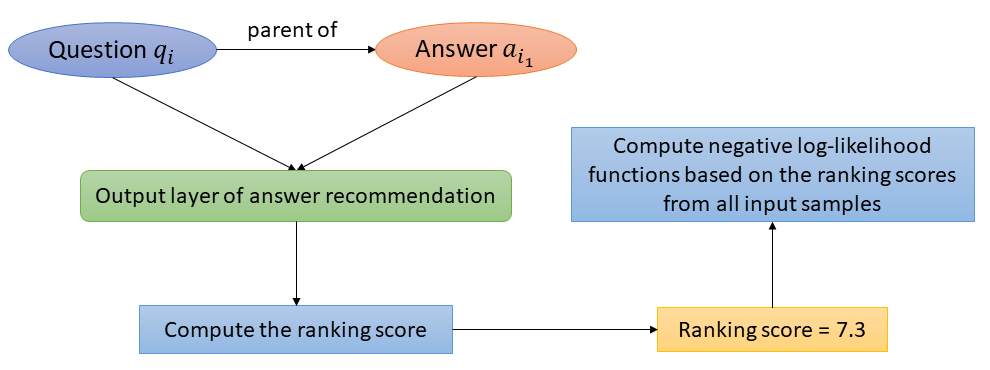}
            \caption{Constraint 2 step (3), (4): compute the ranking score(s) of answers whose owner user(s) has the highest logit, then compute negative log-likelihoods which are averaged to get the overall constraint loss.}
            \label{fig:constraint_2_example_2}
    \end{subfigure}
    \caption{Examples of computing the proposed two cross-task constraints' loss functions. Figures (a)-(b) illustrate the 1st constraint, while figures (c)-(d) illustrate the 2nd constraint.}
\end{figure*}

The constraint of assumption $1$ encourages answer(s) predicted to have the highest score attribute in each sample to rank highly in the recommendation list. Figure (\ref{fig:constraint_1_example_1})-(\ref{fig:constraint_1_example_2}) illustrate an example of computing this constraint loss. The general idea is: $(1)$ For each input sample (i.e., a question with all of its answers), use the output layer of \textbf{answer score classification} to compute all answers' logits; $(2)$ Use the output layer of \textbf{answer recommendation} to get the ranking score(s) of answer(s) with the highest logit; $(3)$ Compute negative log-likelihood functions based on these ranking scores from all input samples, which are then averaged to get the overall constraint loss function $C_1$.

Specifically, we store a sample list $\mathcal{I}_{1}$ before training, where each sample is a question index with a list of the indices of the question's answers. Let the $i$-th sample be $(Q_i, \mathcal{A}(Q_i))$ where $Q_i$ is the question index and $\mathcal{A}(Q_i)$ is the answer indices list. This sample's loss term is calculated as follows: $(1)$ Retrieve the corresponding node embeddings by $Q_i$ and $\mathcal{A}(Q_i)$; $(2)$ Calculate the logit values of the answers associated with $\mathcal{A}(Q_i)$ by Equation (\ref{eq:clf_logit}), where the weight and bias come from the TSOL of answer score classification, and the answer embeddings are those retrieved in Step $(1)$; $(3)$ Create a list $\mathcal{I}_{1}^{'}$ containing the indices of answer(s) with the highest logit value in Step $(2)$; $(4)$ Concatenate the retrieved question embedding with the retrieved embedding of every answer whose index is in $\mathcal{I}_{1}^{'}$. Denote the concatenated embedding(s) as $\mathbf{M_{1}^{i}} \in \R ^{ 2k \times |\mathcal{I}_{1}^{'}| }$; $(5)$ Obtain the ranking score of every answer whose index is in $\mathcal{I}_{1}^{'}$ via Equation (\ref{eq:ranking score}), where the weight, bias, and activation function are from the TSOL of answer recommendation, and the concatenated embeddings are those in $\mathbf{M_{1}^{i}}$; $(6)$ Create a list $\mathcal{R}_{1}^{'}$ containing the ranking scores in Step $(5)$, and then compute the loss term $C_{1}^{i}$ as:
\begin{align}\label{eq:cl term}
    C_{1}^{i} = - \frac{1}{|\mathcal{R}_{1}^{'}|} \sum_{j=1}^{|\mathcal{R}_{1}^{'}|} [\log(f(\mathcal{R}_{1}^{'}[j]))] .
\end{align}
The first constraint loss $C_1$ is the average of loss terms of all samples in $\mathcal{I}_{1}$.
Since minimizing Equation (\ref{eq:cl term}) amounts to maximizing the ranking scores of answers corresponding to $\mathcal{I}_{1}^{'}$, reducing $C_1$ will encourage answer(s) which is predicted to have the highest value of \textit{score} attribute in each sample to rank highly, which enhances the consistency between the answer score classification and answer recommendation tasks.

The constraint of assumption $2$ encourages the answer(s) whose user is predicted to have the highest reputation in each sample to have higher rank in the recommendation list. Figure (\ref{fig:constraint_2_example_1})-(\ref{fig:constraint_2_example_2}) illustrate an example of computing this constraint loss. The general idea is: $(1)$ For each input sample (i.e., a question with all of its answers), find the answers' owner users; $(2)$ Use the output layer of \textbf{user classification} to compute the logits of all these users; $(3)$ Use the output layer of \textbf{answer recommendation} to get the ranking score(s) of answer(s) whose owner(s) has the highest logit; $(4)$ Compute negative log-likelihood functions based on these ranking scores from all samples as in Equation \ref{eq:cl2 term} (the notations are similar to those in Equation \ref{eq:cl term} for the first constraint), which are then averaged to get the overall constraint loss $C_2$.
\begin{align}\label{eq:cl2 term}
    C_{2}^{i} = - \frac{1}{|\mathcal{R}_{2}^{'}|} \sum_{j=1}^{|\mathcal{R}_{2}^{'}|} [\log(f(\mathcal{R}_{2}^{'}[j]))].
\end{align}
Because minimizing Equation (\ref{eq:cl2 term}) is equivalent to maximizing the ranking scores of answers corresponding to $\mathcal{I}_{2}^{'}$, decreasing $C_2$ will provide the answer(s) whose user is predicted to have the highest reputation in each sample with more chance of having a high rank in the recommendation list, which strengthens the consistency between the user reputation classification and answer recommendation tasks.

Admittedly, this approach of designing constraints is somewhat ad-hoc. Nonetheless, these cross-task constraints, which reflect human's domain knowledge of the tasks, not only are lucid and easy to implement but also significantly boost our model's performance as verified in the experiments.

\begin{figure}[t]
    \centering
    \resizebox{\columnwidth}{120pt}{%
    \includegraphics{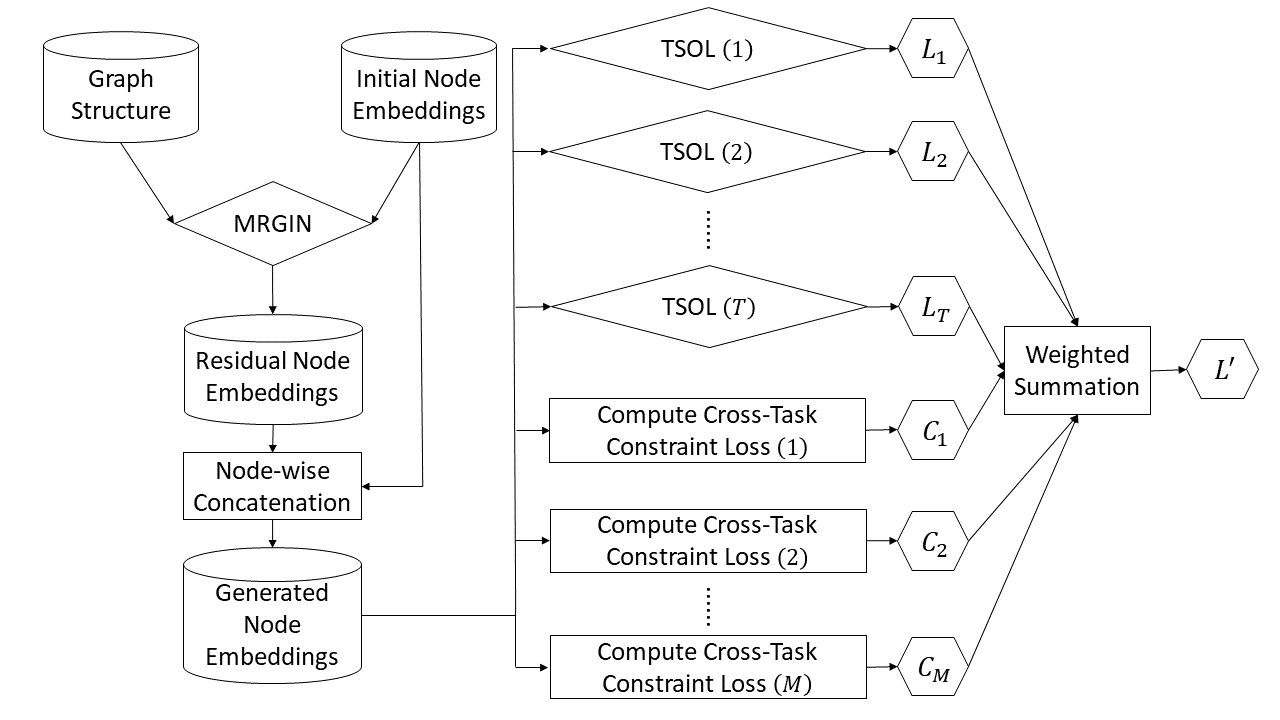}
    }
    \caption{Generic overview of HMTGIN. \textit{TSOL $(i)$} denotes the task-specific output layer of $i$-th task.}
    \label{fig:model}
\end{figure}

\subsection{Heterogeneous Multi-Task Graph Isomorphism Network (HMTGIN)}\label{sec:HMTGIN} 

Finally, we illustrate our MTL algorithm called HMTGIN. 
Suppose the input multi-relational CQA graph consists of graph structure and initial node embeddings, and there are $T$ tasks and $M$ cross-task constraints. The training process of HMTGIN is as follows: $(1)$ The MRGIN model defined in Section \ref{sec:MRGIN} learns new node embeddings by taking in the initial node embeddings and then performing neighborhood aggregations based on the graph structure. The new node embeddings are termed \textit{residual node embeddings}; $(2)$ Skip-connection technique is adopted by performing node-wise concatenation between the initial node embeddings and the residual node embeddings. We call these concatenated embeddings as \textit{generated node embeddings}; $(3)$ The generated node embeddings are shared by every TSOL to calculate corresponding loss $L_{t} (t \in \{1, 2, ..., T\})$ as described in Section \ref{sec:tsol}; $(4)$ Cross-task constraint losses $C_{1}, C_{2}, ...,  C_{M}$ are computed using respective generated node embeddings as described in Section \ref{sec:cross-task constraint}; $(5)$ Compute the total loss $L'$ as: 
    \begin{align}
        L' = \frac{1}{T} \sum_{t=1}^{T} \alpha_t L_{t} + \sum_{i=1}^{M} \beta_i C_{i} ,
    \end{align}
where $\alpha_t (t \in \{1, 2, ..., T\})$ and $\beta_i (i \in \{1, 2, ..., M\})$  are pre-defined constant real numbers.
$(6)$ Update HMTGIN's learnable parameters by back-propagation with respect to the total loss$L'$; $(7)$ Repeat Step $(1)$ - $(6)$ until the termination condition is met.
In evaluation, the generated node embeddings after the last training epoch are shared across different TSOLs to make corresponding predictions. A generic overview of our model is shown in Figure \ref{fig:model} .

\section{Experiments}\label{sec:experiments} 
In this section, we present our experiments.
%we first provide an overview of the Stack Overflow dataset in Section \ref{sec:dataset}. Afterward, we briefly introduce all baselines in Section \ref{sec:baselines}. Thereafter, we describe the experimental settings in Section \ref{sec:settings}. Then, we compare the performance of HMTGIN and baselines in Section \ref{sec:performance_comparison}. Finally, we conduct ablation studies in Section \ref{sec:ablation}.

\subsection{Dataset}\label{sec:dataset}

\begin{figure}[t]
    \centering
    \resizebox{\columnwidth}{120pt}{
    \includegraphics{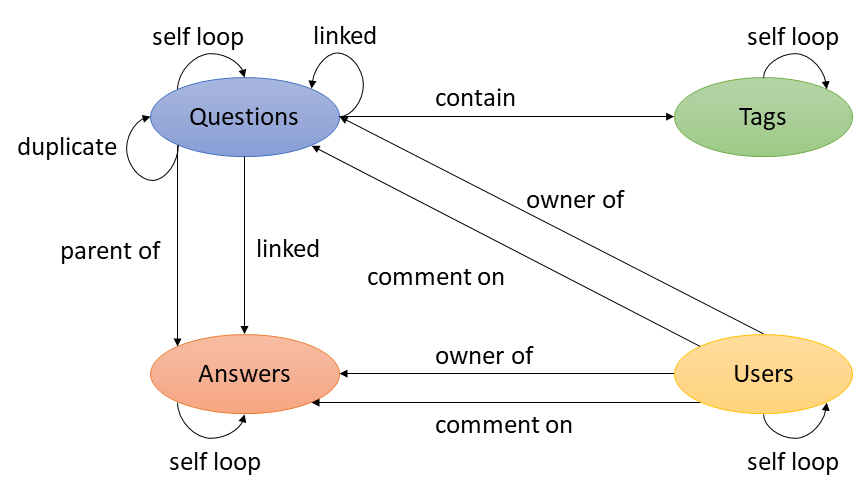}}
    \caption{Schema of the Stack Overflow dataset.}
    \label{fig:SO_Dataset}
\end{figure}

We obtained the Stack Overflow raw dataset from the Stack Exchange Data Dump\footnote{https://archive.org/details/stackexchange}. We then performed data cleaning steps like removing attributes with too many missing values. Afterward, we constructed a million-scale multi-relational graph consisting of questions with at least eight answers, all other types of nodes that are either directly or indirectly connected to these questions, and associated edges. The graph has $4$ node types and $22$ edge types including self loop edge types for all node types and all reverse edge types\footnote{Any two edges which have either different source node types or different destination node types are considered to be of different edge types (e.g., `owner\_of' edge type between Users and Questions are considered to be different from that between Users and Answers).}. 
\begin{comment}
Notice that multiple types of edges might exist between two types of nodes (e.g., `owner\_of' and `comment\_on' edge types between Users and Questions)
\end{comment}
Thus, in our experiments, we do not compare with ordinary GCN and GIN as doing so will lose a lot of edges.
Each node type's cardinality is included in Table \ref{tab:cardinality}, where the total number of nodes is over two million. Figure \ref{fig:SO_Dataset} depicts the graph's schema\footnote{Reverse edge types are omitted for conciseness.}.
\begin{table}[t]
  \centering
  \caption{Cardinalities of all node types.}
    \begin{tabular}{l|r}
    \toprule
    Type  & \multicolumn{1}{l}{Cardinality} \\
    \midrule
    Questions  & 108,113 \\
    \hline
    Answers & 1,212,308 \\
    \hline
    Users & 773,517 \\
    \hline
    Tags  & 55,663 \\
    \bottomrule
    \end{tabular}%
  \label{tab:cardinality}%
\end{table}%

Every node has several attributes (e.g., `Body' attribute of Questions). An attribute would be ignored if it appears in less than $10\%$ of the corresponding type of nodes. Regarding any date-time attribute (e.g., `CreationDate' of Questions), it will be ignored if it is missing in any of the corresponding types of nodes. We performed several text pre-processing operations like expanding contractions on every textual attribute (e.g., `Body' attribute of Answers).
Thereafter, we constructed the initial feature vector for an attribute $A$ as follows: $(1)$ If $A$ is textual, we numericalize it by averaging all its words' pre-trained GloVe \cite{pennington2014glove} embeddings. A vector filled with $0$ is used for missing values; $(2)$ If $A$ is categorical (e.g., `PostTypeId' attribute of Posts of other types),  then we convert it into a one-hot vector. The value of $A$ will be set to $-1$ before conversion if its value is missing; $(3)$ If $A$ is numerical (e.g., `Views' attribute of Users), then we normalize its value into the range $[0, 1]$. The value of $A$ will be set to $0$ before conversion if its value is missing; $(4)$ If $A$ is a date-time attribute, we encode its different aspects of temporal information (e.g., what weekday it is) into a numeric vector.
The initial feature vector of every node is then the concatenation of all its attributes' initial feature vectors.
We applied SVD to reduce the dimension of all nodes' initial feature vectors to $16$ for efficiency. These compressed vectors serve as the initial node embeddings. 
All experiments are only performed on the constructed Stack Overflow dataset as our approach can be directly applied to other CQA platforms with trivial modifications.

\subsection{Baselines}\label{sec:baselines}

We compare our model with the following algorithms: $(1)$ {\bf WDL} \cite{cheng2016wide}; $(2)$ {\bf NCF} \cite{he2017neural}; $(3)$ {\bf CDSSM} \cite{shen2014latent}; $(4)$ {\bf MaLSTM} \cite{mueller2016siamese}; $(5)$ {\bf Text-CNN} \cite{kim2014convolutional} ; $(6)$ {\bf BiLSTM} \cite{schuster1997bidirectional}; $(7)$ {\bf TransE} \cite{BordesUGWY13}; $(8)$ {\bf COMPGCN} \cite{VashishthSNT20}; $(9)$ {\bf KBGAT} \cite{NathaniCSK19}; $(10)$ {\bf BERT (Large)} \cite{DevlinCLT19}. We do not include any of the existing MTL models for CQA in the experiment since none of them can jointly learn the highly heterogeneous tasks targeted in this paper.

\begin{table}[t]
  \centering
  \caption{Baselines and evaluation metrics for all tasks.}
    \begin{tabular}{l|l|l}
    \toprule
    Tasks & Baselines & Metrics \\
    \midrule
    Tag   & WDL, NCF, TransE   & Accuracy, \\
    Recommendation &  COMPGCN, KBGAT & F1 \\
    \hline
    Duplicate Question & CDSSM, MaLSTM, TransE & Accuracy,  \\
    Detection & COMPGCN, KBGAT & F1 \\
    \hline
    Answer  & WDL, NCF & HR@3,  \\
    Recommendation & COMPGCN & NDCG@3 \\
     \hline
    Answer Score & Text-CNN, BiLSTM & Accuracy,  \\
    Classification & COMPGCN, BERT & Macro F1 \\
     \hline
    User Reputation & Text-CNN, BiLSTM & Accuracy,  \\
    Classification & COMPGCN, BERT & Macro F1 \\
  
    \bottomrule
    \end{tabular}%
  \label{tab:baselines_and_metrics}%
\end{table}%

Baseline models for the ranking task are extended to incorporate the loss function of RankNet \cite{burges2005learning} to boost performance. Besides the above baselines, to verify the effectiveness of MRGIN, we also compare HMTGIN with one of its variants termed Heterogeneous Multi-Task Graph Convolutional Networks ({\bf HMTGCN}) which is the same as HMTGIN except that it replaces the MRGIN component of HMTGIN with RGCNs.
The baseline algorithms and evaluation metrics for all tasks are included in Table \ref{tab:baselines_and_metrics}, where F1 means F1 score, HR@3 means Hit Ratio in top-3 list, NDCG@3 means Normalized Discounted Cumulative Gain in top-3 list, and Macro F1 means macro F1 score for multi-class classification\footnote{We omit micro F1 score as it always has the same value as accuracy does in this setting.}. The word embeddings are fixed during training for CDSSM, MaLSTM, Text-CNN, BiLSTM, and BERT.

\subsection{Settings}\label{sec:settings}

We split each task's dataset into training, development, and test sets with an $8:1:1$ ratio. The best hyper-parameter configuration of every model is determined by evaluating the trained models on corresponding development set(s), where the configuration with the best average score over all relevant evaluation metrics is chosen. Multi-task models are compared by their average scores over all respective evaluation metrics on the development sets of all corresponding tasks. Afterward, the trained model with the chosen configuration is examined on the test set(s) to get its test scores for all relevant evaluation metrics. Every instance of a baseline model is trained and evaluated on only one task, whereas every instance of an MTL model is trained and evaluated on all tasks.

The hyper-parameters we have tuned for HMTGIN as follows, where the values used in the best configuration of HMTGIN are shown in bold.
\begin{enumerate}
    \item number of hidden layer: $\mathbf{0}, 1, 2, 3$;
    \item hidden dimension(s) : $8, \mathbf{16}, 32, 64$;
    \item whether the $\epsilon$ coefficient in Equation \ref{eq:MRGIN} is trainable or not: yes, \textbf{no};
    \item dropout rate in all MRGIN layer: $\mathbf{0}, 0.1, 0.2, 0.3$;
    \item number of mlp layer in each MRGIN layer: $0, \mathbf{1}, 2, 3, 4$.
\end{enumerate}
We train each HMTGIN instance for $135$ epochs, where the checkpoint is saved whenever the average score increases. The coefficients of both cross-task constraints are $1$. The coefficients associated with the losses of both duplication question detection and answer score classification are $7$, whereas the others are $1$. The model is trained using Adam \cite{kingma2014adam} algorithm in full batch. The $\sigma$ activation function in Equation \ref{eq:MRGIN} and Equation \ref{eq:ranking score} is Leaky ReLU. The value of the $\epsilon$ coefficient in Equation \ref{eq:MRGIN} is $0$. The learning rate is reduced by $0.5$ in every $50$ epochs.

\begin{comment}
\begin{table*}[htbp]
  \centering
  \caption{Performance of all baseline models and HMTGIN.}
  \resizebox{2.1\columnwidth}{60 pt}{%
    \begin{tabular}{|c|c|c|c|c|c|c|c|c|c|c|}
    \toprule
    Models  Tasks Metrics & TR Accuracy & TR F1 & DQD Accuracy & DQD F1 & AR HR@3 & AR NDCG@3 & ASC Accuracy & ASC Macro F1 & URC Accuracy & URC Macro F1 \\
    \midrule
    WDL   & 0.87  & 0.786 & N/A   & N/A   & 0.624 & 0.49  & N/A   & N/A   & N/A   & N/A \\
    \midrule
    NCF   & 0.871 & 0.786 & N/A   & N/A   & 0.631 & 0.498 & N/A   & N/A   & N/A   & N/A \\
    \midrule
    CDSSM & N/A   & N/A   & 0.795 & 0.656 & N/A   & N/A   & N/A   & N/A   & N/A   & N/A \\
    \midrule
    MaLSTM & N/A   & N/A   & 0.794 & 0.601 & N/A   & N/A   & N/A   & N/A   & N/A   & N/A \\
    \midrule
    Text-CNN & N/A   & N/A   & N/A   & N/A   & N/A   & N/A   & 0.421 & 0.423 & 0.395 & 0.393 \\
    \midrule
    BiLSTM & N/A   & N/A   & N/A   & N/A   & N/A   & N/A   & 0.435 & 0.438 & 0.405 & 0.396 \\
    \midrule
    HMTGCN & 0.911 & 0.866 & 0.773 & 0.645 & 0.717 & 0.589 & 0.434 & 0.429 & 0.423 & 0.417 \\
    \midrule
    \midrule
    HMTGIN & \textbf{0.921} & \textbf{0.878} & \textbf{0.803} & \textbf{0.685} & \textbf{0.737} & \textbf{0.619} & \textbf{0.444} & \textbf{0.439} & \textbf{0.483} & \textbf{0.47} \\
    \bottomrule
    \end{tabular}%
  }
  \label{tab:performance}%
\end{table*}%

\end{comment}

\subsection{Performance Comparison}\label{sec:performance_comparison}

We abbreviate task names as follows: `{\bf TR}' means tag recommendation; `{\bf DQD}' denotes duplicate question detection; `{\bf AR}' means answer recommendation; `{\bf ASC}' denotes answer score classification; `{\bf URC}' means user reputation classification. Regarding each task, we train and test the corresponding model instances with their respective best hyper-parameters under three different random seeds. The mean test scores and the standard deviations of all models are shown in Table \ref{tab:pc_tag_rec}, \ref{tab:pc_dqd}, \ref{tab:pc_ar}, \ref{tab:pc_asc}, and \ref{tab:pc_urc} respectively \footnote{The numbers to the left of `$\pm$' symbols are the mean scores and the numbers to the right of `$\pm$' symbols are the corresponding standard deviations}. The best mean score in each task for every corresponding evaluation metric is marked in bold.

\begin{table}[t]
  \centering
  \caption{Performance comparison in the TR task.}
    \begin{tabular}{c|cc}
    \toprule
    \backslashbox{Models}{Metrics} & Accuracy & F1 \\
    \midrule
    WDL   & 0.870 $\pm$ 0.002  & 0.786 $\pm$ 0.001\\
    NCF   & 0.871 $\pm$ 0.001 & 0.786 $\pm$ 0.001 \\
    TransE   & 0.879 $\pm$ 0.000 & 0.795 $\pm$ 0.000 \\
    COMPGCN   & 0.895 $\pm$ 0.002 & 0.810 $\pm$ 0.001 \\
    KBGAT   & 0.882 $\pm$ 0.001 & 0.803 $\pm$ 0.001 \\
    \hline
    HMTGCN & 0.911 $\pm$ 0.002 & 0.866 $\pm$ 0.002\\
    \hline
    HMTGIN & \textbf{0.921} $\pm$ 0.000& \textbf{0.878} $\pm$ 0.000 \\
    \bottomrule
    \end{tabular}%
  \label{tab:pc_tag_rec}%
\end{table}%

\begin{table}[t]
  \centering
  \caption{Performance comparison in the DQD task.}
    \begin{tabular}{c|cc}
    \toprule
    \backslashbox{Models}{Metrics} & Accuracy & F1 \\
    \midrule
    CDSSM & 0.795 $\pm$ 0.004 & 0.656 $\pm$ 0.003 \\
    MaLSTM & 0.794 $\pm$ 0.003 & 0.601 $\pm$ 0.002 \\
    TransE   & 0.683 $\pm$ 0.001 & 0.552 $\pm$ 0.001 \\
    COMPGCN   & 0.702 $\pm$ 0.003 & 0.583 $\pm$ 0.002 \\
    KBGAT   & 0.721 $\pm$ 0.002 & 0.591 $\pm$ 0.002 \\
    \hline
    HMTGCN & 0.773 $\pm$ 0.001 & 0.645 $\pm$ 0.002 \\
    \hline
    HMTGIN & \textbf{0.803} $\pm$ 0.001 & \textbf{0.685} $\pm$ 0.001 \\
    \bottomrule
    \end{tabular}%
  \label{tab:pc_dqd}%
\end{table}%

\begin{table}[t]
  \centering
  \caption{Performance comparison in the AR task.}
    \begin{tabular}{c|cc}
    \toprule
    \backslashbox{Models}{Metrics} & HR@3 & NDCG@3 \\
    \midrule
    WDL   & 0.624 $\pm$ 0.001 & 0.49 $\pm$ 0.001 \\
    NCF   & 0.631 $\pm$ 0.001 & 0.498 $\pm$ 0.001 \\
    COMPGCN   & 0.697 $\pm$ 0.001 & 0.583 $\pm$ 0.002 \\
    \hline
    HMTGCN & 0.717 $\pm$ 0.002 & 0.589 $\pm$ 0.003 \\
    \hline
    HMTGIN & \textbf{0.737} $\pm$ 0.000 & \textbf{0.619} $\pm$ 0.000 \\

    \bottomrule
    \end{tabular}%
  \label{tab:pc_ar}%
\end{table}%

\begin{table}[t]
  \centering
  \caption{Performance comparison in the ASC task.}
    \begin{tabular}{c|cc}
    \toprule
    \backslashbox{Models}{Metrics} & Accuracy & Macro F1 \\
    \midrule
    Text-CNN & 0.421 $\pm$ 0.001 & 0.423 $\pm$ 0.002 \\
    BiLSTM & 0.435 $\pm$ 0.004 & 0.438 $\pm$ 0.005 \\
    COMPGCN   & 0.331 $\pm$ 0.002 & 0.342 $\pm$ 0.001 \\
    BERT   & 0.442 $\pm$ 0.001 & 0.438 $\pm$ 0.001 \\
    \hline
    HMTGCN & 0.434 $\pm$ 0.002 & 0.429 $\pm$ 0.002 \\
    \hline
    HMTGIN & \textbf{0.444} $\pm$ 0.001 & \textbf{0.439} $\pm$ 0.002 \\
    \bottomrule
    \end{tabular}%
  \label{tab:pc_asc}%
\end{table}%

\begin{table}[t]
  \centering
  \caption{Performance comparison in the URC task.}
    \begin{tabular}{c|cc}
    \toprule
    \backslashbox{Models}{Metrics} & Accuracy & Macro F1 \\
    \midrule
    Text-CNN & 0.395 $\pm$ 0.003 & 0.393 $\pm$ 0.003\\
    BiLSTM & 0.405 $\pm$ 0.001 & 0.396 $\pm$ 0.003 \\
    COMPGCN   & 0.314 $\pm$ 0.001 & 0.302 $\pm$ 0.001 \\
    BERT & 0.459 $\pm$ 0.001 & 0.436 $\pm$ 0.001 \\
    \hline
    HMTGCN & 0.423$\pm$ 0.002 & 0.417 $\pm$ 0.001 \\
    \hline
    HMTGIN & \textbf{0.483} $\pm$ 0.002& \textbf{0.470} $\pm$ 0.001\\
    \bottomrule
    \end{tabular}%
  \label{tab:pc_urc}%
\end{table}%

In summary, HMTGIN outperforms all baselines in all tasks, where the improvements are up to $16.9\%$, which distinctly demonstrates its effectiveness.

Regarding link prediction, the improvements in accuracy are up to $12\%$, and the improvements in F1 score are up to $13.3\%$. 
%
\begin{comment}
Unlike in TR task, where HMTGIN outruns the baselines by a large margin, the baselines and HMTGIN perform in a similarly unsatisfactory level in DQD task. This is probably because DQD task is very challenging in the Stack Overflow dataset constructed in this paper as there are only 2465 duplicate question pairs available in the dataset, which results in a data scarcity issue.
\end{comment}
%
Regarding ranking, the improvements are up to $11.3\%$ and $12.9\%$ in HR@3 and NDCG@3 respectively. Such considerable improvements indicate that the proposed cross-task constraints are significant.
Regarding classification, all HMTGIN's scores in URC task substantially exceed those of the corresponding best baselines (the improvements over accuracy and macro F1 are $2.4\%$ and $3.4\%$ respectively), whereas in ASC task, HMTGIN only slightly outperforms the corresponding best baselines. This might result from that the best baseline is already powerful enough for ASC task, which leads to a small room for further improvement.
By comparing the performance of HMTGIN and HMTGCN, we can see that HMTGIN consistently outperforms HMTGCN in all tasks, where the improvements are up to $6\%$, and in two out of the five tasks, the improvements are at least $3\%$. This result shows that GIN is more powerful than GCN.

\subsection{Ablation Studies}\label{sec:ablation}

We conduct the following ablation studies: $(1)$ comparing Single Task Learning (STL) settings with MTL settings; $(2)$ examining the influence of the cross-task constraints. $(3)$ parameter sensitivity study on the number of MLP layers in each MRGIN layer. Every numerical result is the average over three random seeds.

\subsubsection{STL Settings VS MTL Settings}

\begin{figure}[t]
    \centering
    \includegraphics[width=0.49\textwidth,height=0.11\textwidth]{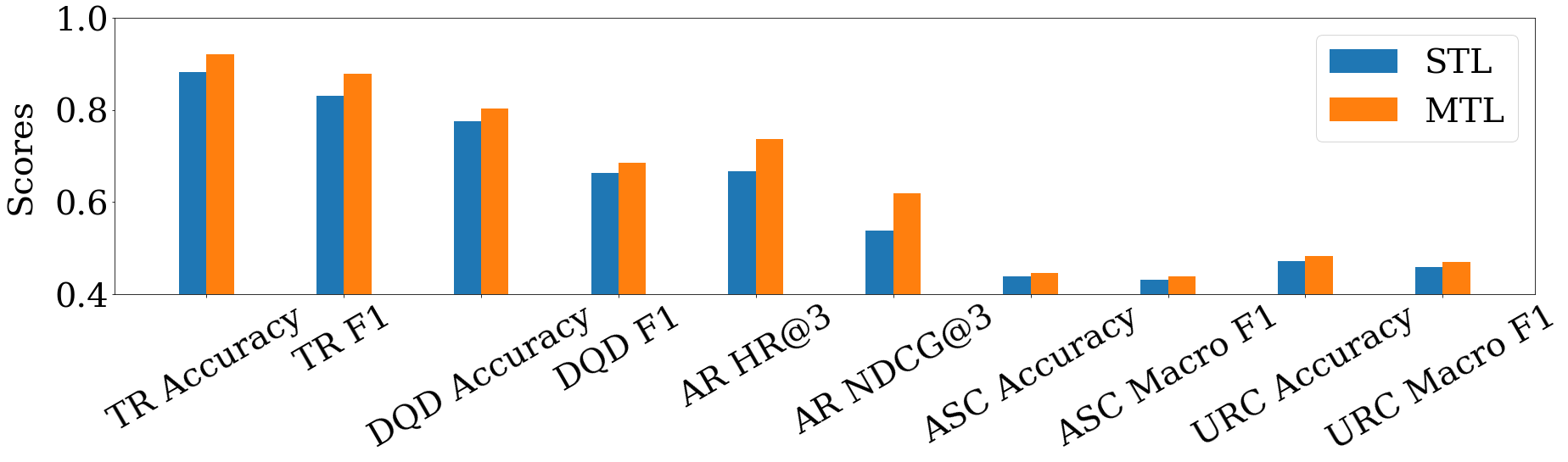}
    \caption{Comparison between STL and MTL settings of HMTGIN.}
    \label{fig:stl_vs_mtl}
\end{figure}

 The STL scores together with the MTL ones are shown in Figure \ref{fig:stl_vs_mtl}, and the x-axis contains the task names and respective metrics.

Generally speaking, the MTL instance outperforms all of its STL counterparts for all evaluation metrics (i.e., there is no negative transfer), where the improvements are up to $8\%$, which justifies the effectiveness of our MTL strategy. 
In particular, the MTL instance surpasses the STL instances by a large margin in TR ($3.8\%$ and $4.7\%$ in accuracy and F1 score respectively), DQD ($2.8\%$ and $2.2\%$ in accuracy and F1 score respectively), and AR ($7.1\%$ and $8\%$ in HR@3 and NDCG@3 respectively). By contrast, the performance gain in the ASC task ($0.7\%$ in both accuracy and macro F1 score) and URC task ($1.2\%$ and $1.1\%$ in accuracy and macro F1 score respectively) is not very noticeable. Such a distinction indicates that learning signals from the MTL mechanism are more crucial for the first three tasks than the last two tasks.

\subsubsection{Role of Cross-task Constraints}

\begin{figure}[t]
    \centering
    \includegraphics[width=0.45\textwidth,height=0.2\textwidth ]{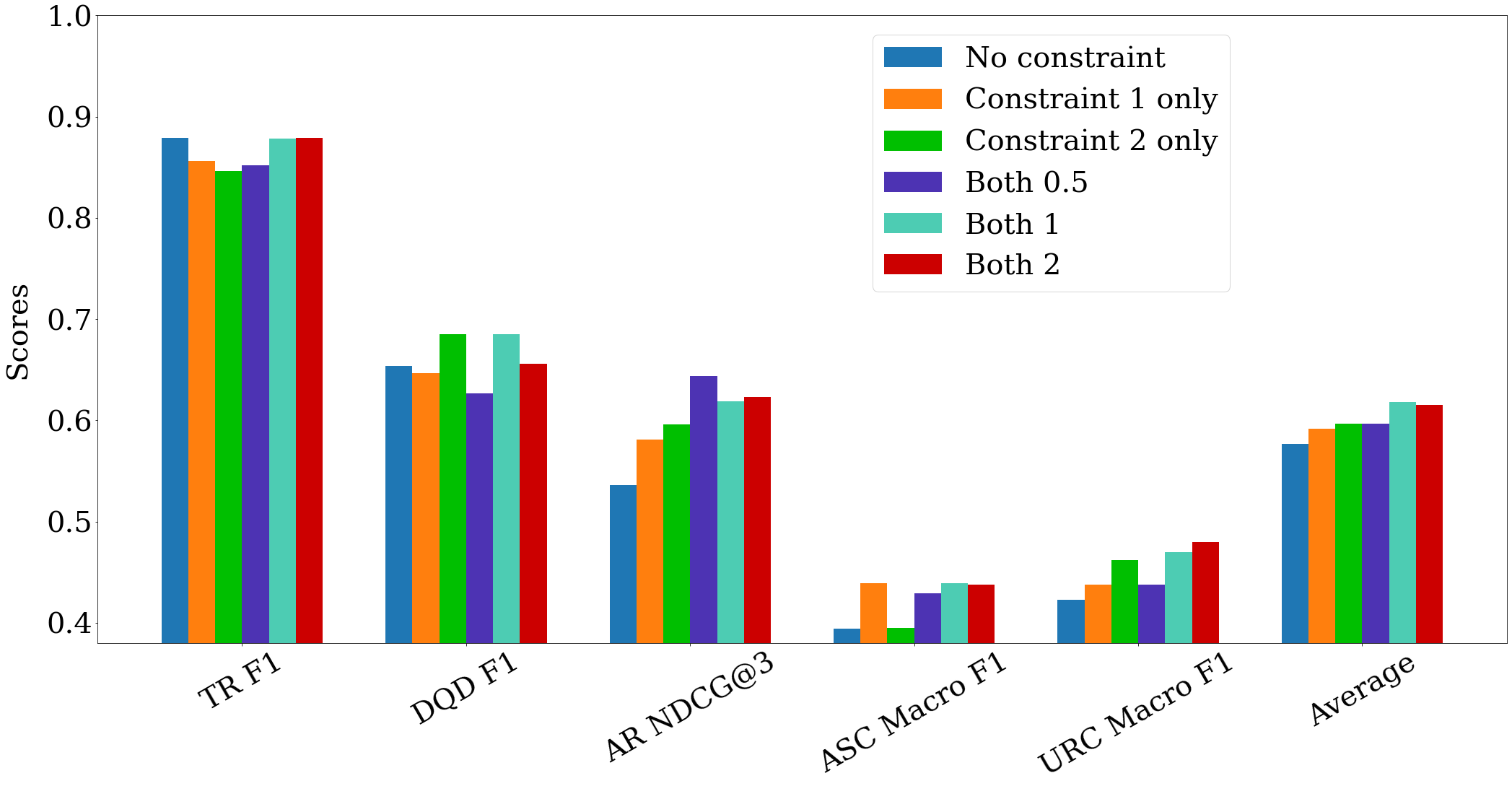}
    \caption{Comparison between different settings of HMTGIN's cross-task constraints.}
    \label{fig:constraints_experiments}
\end{figure}

We conduct experiments on the following settings of cross-task constraints to examine their significance, where the results are included in Figure \ref{fig:constraints_experiments}: $(1)$ \textit{No constraint} means removing both constraints described in Section \ref{sec:cross-task constraint}; $(2)$ \textit{Constraint 1 only} denotes retaining only constraint 1, and similarly for \textit{Constraint 2 only}; $(3)$ \textit{Both 0.5} denotes keeping both constraints, and setting both their coefficients as $0.5$, and similarly for \textit{Both 1}\footnote{Configuration of the HMTGIN instance in Section \ref{sec:performance_comparison} corresponds to this setting.} and  \textit{Both 2}. We only show and discuss the F1 scores for link prediction tasks, NDCG@3 score for ranking task, and macro F1 scores for classification task here due to space limitation, and the pattern of the scores which are not included here is almost the same. The x-axis is similar to that of Figure \ref{fig:stl_vs_mtl}, where the `Average' includes the average value over all the five shown scores of the instance under each of the above five settings.

In general, \textit{Both 1} instance achieves the best average score, which shows that the proposed cross-task constraints are beneficial to the overall performance. Moreover, keeping both constraints almost always improves performance in the tasks related to the constraints (i.e., AR, ASC, and URC), except in a few cases where there are still some settings that retain both constraints and have the best scores for the corresponding evaluation metrics. For instance, \textit{Both 0.5}, \textit{Both 1}, and \textit{Both 2} instances all outperform the remaining settings in both evaluation metrics for AR, where the improvements for NDCG@3 range from $2.3\%$ to $25\%$ respectively. These enhancements are attributed to the cross-task constraints which strengthen the consistency between related tasks, as illustrated in Section \ref{sec:cross-task constraint}. 
Specifically, \textit{No constraint} and \textit{Constraint 1 only} instances consistently perform worse than the setting \textit{Constraint 2 only} in URC (the average difference for macro F1 is $3.2\%$), which empirically justifies the analyses in Section \ref{sec:cross-task constraint} as the constraint 2 is established on the relationship between URC and AR. In addition, \textit{Constraint 1 only} instance surpasses both \textit{Constraint 2 only} and \textit{No constraint} instances in ASC (the average improvements in macro F1 is $8.9\%$), which is also consistent with the explanation in Section \ref{sec:cross-task constraint} since the constraint 1 reinforces the coherence between AR and ASC. Another interesting observation is that the average score of \textit{Both 0.5} instance is quite close to those of \textit{Constraint 1 only}, \textit{Constraint 2 only} and \textit{No Constraint} instances, where the improvements range from $0.1\%$  to $2.1\%$. This might be because the small weights of both coefficients weaken the regularization power of the constraints. However, the average score of the \textit{Constraint 2 only} instance is worse than that of the \textit{Constraint 1 only} instance, although the gap is tiny ($0.3\%$). This is probably because the high weights of both coefficients somewhat impede the optimization of losses. Accordingly, choosing appropriate values of both constraints' coefficients is critical.

\subsubsection{Parameter sensitivity study on the number of MLP layers in each MRGIN layer}

\begin{figure}
    \centering
    \resizebox{\columnwidth}{100pt}{%
    \includegraphics{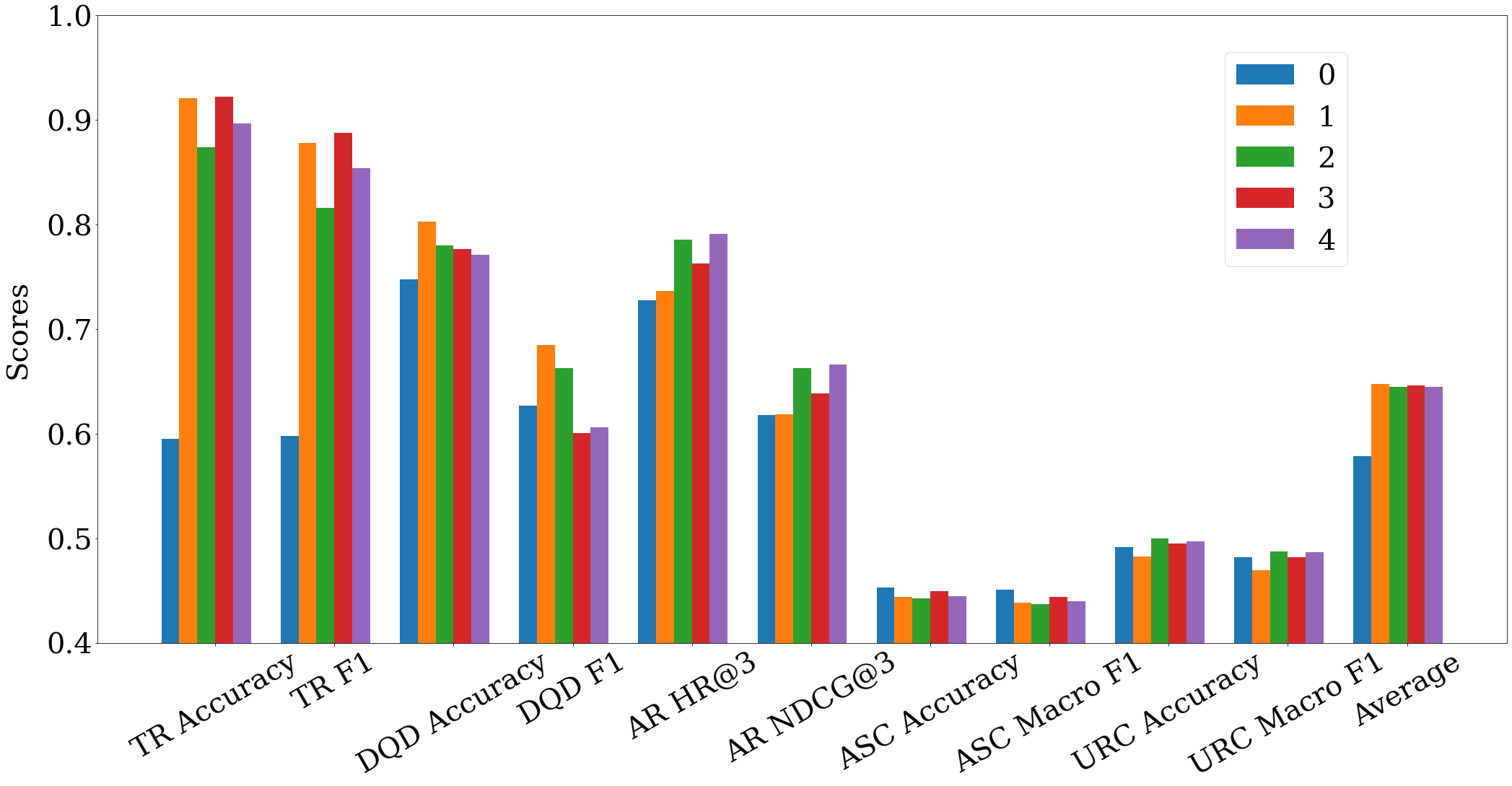}
    }%
    \caption{Comparison between different numbers of the MLP layer in each MRGIN layer.}
    \label{fig:num_mlp_layer}
\end{figure}

The performance with different number of MLP layer in each MRGIN layer is shown in Figure \ref{fig:num_mlp_layer}, where the \textit{0} is the instance with no MLP module , the \textit{1} is the instance with 1 MLP layer, and so on\footnote{The \textit{1} instance corresponds to the configuration in Section \ref{sec:performance_comparison}.}.

To summarize, the presence of MLP layer greatly boosts the performance (the average improvements are up to $6.9\%$), which can be attributed to the universal approximation theorem \cite{hornik1991approximation} ensuring that MLP can learn the composition of the multiset functions that allows GIN to have the maximum discriminative power among all GNN variants \cite{xu2018powerful}. 
Furthermore, the instance with one MLP layer has the best average score (i.e., $64.8\%$), even though the other instances with MLP module are only slightly weaker than it (the largest margin in the average score is only $0.3\%$). This indicates that altering the number of MLP layer in each MRGIN layer of HMTGIN have limited effect when there is at least one MLP layer in each MRGIN layer.  Moreover, increasing the number of MLP layer does not exhibit any generally monotonic trend in the average scores (the changes from zero MLP layer to one MLP layer, from one to two MLP layers, from two MLP layers to three MLP layers, and from three MLP layers to four MLP layers are $6.9\%$, $-0.3\%$, $0.1\%$ and $- 0.1\%$ respectively). 
The performance gain offered by the MLP module is most distinct in TR, where the improvements from the instance without MLP layer to instances with MLP layer are up to $32.7\%$ and $29.0\%$ for accuracy and F1 respectively. Such considerable improvements might be since the MLP layer is highly effective in capturing the tags' semantics. By contrast, the instances with MLP module tend to perform worse compared with those without MLP module in classification tasks, especially in ASC, where the instance without MLP layer outperforms all the instances with MLP layer in both evaluation metrics (the score differences range from $0.3\%$ to $1.0\%$ and $0.7\%$ to $1.4\%$ for accuracy and macro F1 respectively). Such results suggest that MLP is not very helpful for distinguishing the data characteristics in MTL.

\section{Conclusion and Future Work}
In this paper, we propose a multi-relational graph based MTL model named HMTGIN which efficiently tackles CQA with considerable task and graph heterogeneity. To the best of our knowledge, HMTGIN is the first MTL model solving CQA tasks from the angle of multi-relational graphs. To evaluate the effectiveness of HMTGIN, we build a novel million-scale multi-relational graph CQA dataset. 
Experiments of five tasks demonstrate that HMTGIN surpasses all baselines in all tasks. Further ablation studies manifest the substantial role of the proposed cross-task constraints and MTL strategy.
An interesting future direction would be investigating how to develop a more principled way to incorporate the domain knowledge by imposing cross-task constraints on multi-relational graph based MTL algorithms for CQA tasks. 

\section{Acknowledgement}
The authors of this paper were supported by the NSFC Fund (U20B2053) from the NSFC of China, the RIF (R6020-19 and R6021-20) and the GRF (16211520) from RGC of Hong Kong, the MHKJFS (MHP/001/19) from ITC of Hong Kong, with special thanks to the WeChat-HKUST WHAT Lab on Artificial Intelligence Technology.

\bibliographystyle{ACM-Reference-Format}
\bibliography{Reference}

\end{document}